\documentclass{llncs}
\usepackage[dvips]{graphicx}
\usepackage{algorithm} 
\usepackage{algorithmic}  
\usepackage{latexsym}
\begin{document}

\title{\textbf{\large Clustering with Lattices in the Analysis of Graph Patterns}}
\author{Edgar H. de Graaf \and Joost N. Kok \and Walter A. Kosters}
\institute{\textit{Leiden Institute of Advanced Computer Science\\
Leiden University, The Netherlands} \\
\texttt{edegraaf@liacs.nl} }

\pagestyle{empty}
\maketitle
\thispagestyle{empty}
\begin{abstract}
Mining frequent subgraphs is an area of research where we have a given set of graphs
(each graph can be seen as a transaction), and we search for (connected) subgraphs contained
in many of these graphs.
In this work we will discuss techniques used in our framework \textsc{Lattice2SAR} 
for mining and analysing frequent subgraph data and their corresponding lattice information. 
Lattice information is provided by the graph mining algorithm \textsc{gSpan};
it contains all supergraph-subgraph relations of the frequent subgraph patterns --- and their supports.

\textsc{Lattice2SAR} is in particular used in the analysis of frequent 
graph patterns where the graphs are molecules and the frequent subgraphs are fragments.
In the analysis of fragments one is interested in the molecules where
patterns occur. This data can be very extensive and in this paper we focus
on a technique of making it better available by using the lattice
information
in our clustering. Now we can reduce the number of times
the highly compressed occurrence data needs to be accessed by the user.
The user does not have to browse all the occurrence data in search of patterns
occurring in the same molecules. Instead one can directly see which frequent
subgraphs are of interest.
\end{abstract}

\section{Introduction}

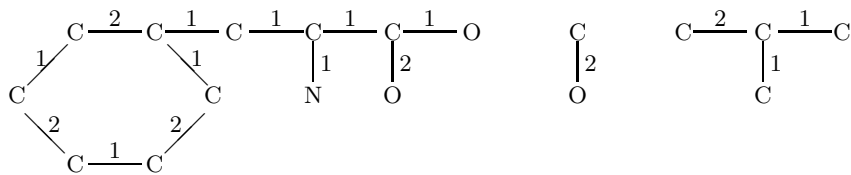
\begin{figure}[!ht]
\begin{center}
\begin{picture}(200,50)(50,10)
%\put(5,10){\line(-1,0){20}}
%\put(-20,10){\makebox(0,0){H}}
%\put(5,60){\line(-1,0){20}}
%\put(-20,60){\makebox(0,0){H}}
\put(10,10){\makebox(0,0){C}}
\put(40,10){\makebox(0,0){C}}
\put(10,60){\makebox(0,0){C}}
\put(40,60){\makebox(0,0){C}}
\put(70,60){\makebox(0,0){C}}
\put(100,60){\makebox(0,0){C}}
\put(54,65){\makebox(0,0){1}}
\put(44,60){\line(1,0){20}}
\put(86,65){\makebox(0,0){1}}
\put(76,60){\line(1,0){20}}
\put(130,60){\makebox(0,0){C}}
\put(160,60){\makebox(0,0){O}}
\put(114,65){\makebox(0,0){1}}
\put(104,60){\line(1,0){20}}
\put(144,65){\makebox(0,0){1}}
\put(134,60){\line(1,0){20}}
%\put(190,60){\makebox(0,0){H}}
%\put(164,60){\line(1,0){20}}

\put(100,36){\makebox(0,0){N}}
\put(105,48){\makebox(0,0){1}}
\put(100,56){\line(0,-1){15}}

\put(130,36){\makebox(0,0){O}}
\put(135,48){\makebox(0,0){2}}
\put(130,56){\line(0,-1){15}}

\put(62,36){\makebox(0,0){C}}
\put(-12,36){\makebox(0,0){C}}
\put(25,15){\makebox(0,0){1}}
\put(15,10){\line(1,0){20}}
\put(25,65){\makebox(0,0){2}}
\put(15,60){\line(1,0){20}}
\put(2,25){\makebox(0,0){2}}
\put(6,14){\line(-1,1){15}}
\put(48,25){\makebox(0,0){2}}
\put(44,14){\line(1,1){15}}
\put(56,50){\makebox(0,0){1}}
\put(60,40){\line(-1,1){15}}
\put(-3,50){\makebox(0,0){1}}
\put(-8,40){\line(1,1){15}}

\put(200,60){\makebox(0,0){C}}
\put(200,36){\makebox(0,0){O}}
\put(205,48){\makebox(0,0){2}}
\put(200,56){\line(0,-1){15}}

\put(240,60){\makebox(0,0){C}}
\put(270,60){\makebox(0,0){C}}
\put(300,60){\makebox(0,0){C}}
\put(254,65){\makebox(0,0){2}}
\put(244,60){\line(1,0){20}}
\put(286,65){\makebox(0,0){1}}
\put(276,60){\line(1,0){20}}
\put(270,36){\makebox(0,0){C}}
\put(275,48){\makebox(0,0){1}}
\put(270,56){\line(0,-1){15}}

\end{picture}
\caption{An example of a possible graph (the amino acid Phenylalanine) in the molecule dataset
and two of its many (connected) subgraphs, also called patterns or fragments.}
\label{fig:subgraph}
\end{center}
\end{figure}

% * Vertel iets generieks over onze oplossing *
Mining frequent patterns is an important area of data mining
where we discover substructures that occur often in (semi-)structured data.
The research in this work will be in the area of frequent subgraph mining.
These \emph{frequent subgraphs} are connected vertex- and edge-labeled graphs that are subgraphs of a 
given set of graphs, traditionally also referred to as \emph{transactions}, 
at least $\mathit{minsupp}$ times. The example of Figure~\ref{fig:subgraph} shows
a graph and two of its subgraphs. 

In this paper we will use results from frequent subgraph mining and visualize 
the frequent subgraphs by means of clustering, where their co-occurrences in the same
transactions are used in the distance measure. Clustering makes it possible to obtain a quicker selection
of the right frequent subgraphs for a more detailed look at their occurrence.

Before explaining what is meant by lattice information we first need to discuss \emph{child-parent}
relations in frequent subgraphs, also known as patterns. Patterns are 
generated by extending smaller patterns with one extra edge. The smaller pattern 
can be called a \emph{parent} of the bigger pattern that it is extended to. If we would draw
all these relations, the drawing would be shaped like a lattice, hence we call this data
\emph{lattice information}.

We further analyse frequent subgraphs and their corresponding lattice information with different 
techniques in our framework \textsc{Lattice2SAR} for mining and analysing frequent subgraph data. 
One of the techniques in this framework is the analysis of graphs in
which frequent subgraphs occur, via clustering. Another important functionality is the browsing of lattice information
from parent to child and from child to parent. In this paper we will present the clustering.

\vspace*{-1cm}

Our working example is the analysis of patterns (fragments) in molecule data, since \textsc{Lattice2SAR}
was originally made to handle molecule data. 
Obviously molecules are stored in the form of graphs, the molecules are the transactions (see Figure~\ref{fig:subgraph} for an example). 
However, the techniques presented here are not particular to molecule data
(we will also not discuss any chemical or biological issues). For example one can extract
user behaviour from access logs of a website. This behaviour can be stored in the form of
graphs and can as such be mined with
the techniques presented here.

\begin{figure}[!ht]
\begin{center}
\includegraphics[width=8.0cm]{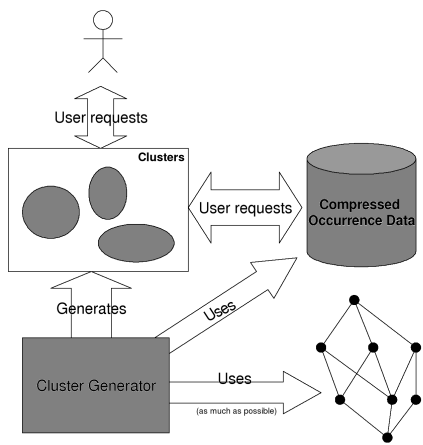}
\caption{Diagram of how the cluster browsing part of \textsc{Lattice2SAR} is used.}
\label{fig:clusteruse}
\vspace*{-10mm}
\end{center}
\end{figure}

The distance between patterns can be measured by calculating in how many graphs (or molecules) only one of the two patterns occurs. If this never 
happens then these patterns are very close to each other. If this always happens then their distance
is very large. In both cases the user is interested to know the reason. In our working example the
chemist might want to know which different patterns seem to occur in the same subgroup of effective medicines
or which patterns occur in different subgroups of effective medicines. In this paper we will
present an approach to solve this problem that uses clustering. Furthermore all occurrences for the frequent subgraphs will
be discovered by a graph mining algorithm and this  occurrence information will be highly compressed before storage. 
Because of this, requesting these occurrences will be costly. Through our method of clustering time will be saved 
if the user uses the clusters to select interesting patterns, see Figure \ref{fig:clusteruse} for an overview.

% * voorbeeld plaatje?

We will define our method of clustering and show its usefulness. To this end, this paper makes the
following contributions:
\\
--- Our first contribution will be that we will \textbf{introduce an algorithm for clustering 
frequent subgraphs} allowing the user to quickly see interesting relations, e.g., subgraphs
occur in the same transactions, and quicker select the right occurrence details from
the compressed storage (all sections and specifically Section~\ref{relative}).
\\
--- We will define a measure of calculating distances between patterns and
\textbf{show how the lattice information can be used for faster calculation}
(Section~\ref{distance} and Section~\ref{group}).
\\
--- The lattice information can be used to make groups of patterns and in this
way \textbf{clarity of the visualization can be improved} due to less points in the 2-dimensional model. 
In Section~\ref{group} we will introduce this preprocessing step.
\\
--- Preprocessing will also make \textbf{faster clustering possible by reducing cluster points},
diminishing requests for occurrence counting (Section~\ref{group} and Section~\ref{exps}).
\\
--- Finally through experiments \textbf{the effectiveness of our clustering is shown}
and the resulting cluster model is analyzed (Section~\ref{exps}).

% * Het werk van andere, related work *
This research is related to research on clustering, in particular
of molecules. Also our work is related to frequent subgraph mining
and frequent pattern mining when lattices are discussed. In \cite{Zaki1}
Zaki et al. discuss different ways for searching through the lattice
and they propose the \textsc{Eclat} algorithm.

Clustering in the area of biology is important because of the visualization that
it can provide. E.g., \cite{SamsonovaEV} Samsonova et al. discuss the use of
Self-Organizing Maps (SOMs) for clustering protein data. In general our work is 
related to SOMs as developed by Kohonen (see \cite{KohonenT}),
in the sense that SOMs are also used to visualize data through
a distance measure. SOMs have been used in a biological context many
times, for example in \cite{HankeJ,MahonyS}. In some cases 
molecules are clustered via numeric data describing each molecule, in
\cite{XuJ} clustering such data is investigated.

Our package of mining techniques for molecules is called \textsc{Lattice2SAR}; it
makes use of a graph miner called \textsc{gSpan}, introduced in \cite{YanX} 
by Yan and Han. This implementation generates the patterns organized 
as a lattice and a separate compressed file of occurrences
of the patterns in the graph set (molecules).

In this work a method of pushing and pulling points in accordance with a distance measure is used.
This technique was used before by Cocx et al. in \cite{CocxT} to cluster criminal careers and was 
developed in \cite{KostersWA}. This method of clustering was chosen since we only know the distance between
two patterns. We don't know the precise $x$ and $y$ coordinates of the patterns, so we can not use
standard methods of discovering clusters, e.g., K-Means (see \cite{MacQueenJB}). 
The algorithm from \cite{KostersWA} is different from 
clustering with self-organizing maps as SOMs adapt the weight vector of each neuron toward an input 
vector. In our problem no such input vector exists and each point in the cluster is linked with one
graph or one group of graphs.

\section{Distance Measure}\label{distance}

For any clustering algorithm you have to at least know the distance between the points in the model.
As was mentioned in the introduction, we are interested to know if patterns occur in the same 
graphs in the dataset of graphs. Patterns in this work are \emph{connected frequent subgraphs} where
all vertices of the subgraph can be found in $\mathit{minsupp}$ graphs of the dataset 
with matching labels and connections between the vertices (see Figure \ref{fig:subgraph} for an example). 
If a subgraph occurs at different positions in a graph, it is counted only once.
Here $\mathit{minsupp}$ is a user-defined threshold for frequency.

The distance measure will compute how often frequent subgraphs occur in the same graphs of the dataset.
In the case of our working example it will show if different fragments (frequent subgraphs) exist
in the same molecules. Formally we will define the distance measure 
in the following way (for graphs $g_{1}$ and $g_{2}$):

\begin{equation}\label{one} 
\mathit{dist}(g_{1}, g_{2}) = \frac{\mathit{support}(g_{1}) + \mathit{support}(g_{2}) - 2 \cdot \mathit{support}(g_{1} \wedge g_{2})}
{\mathit{support}(g_{1} \vee g_{2})}
\end{equation}

\noindent Here $\mathit{support}(g)$ is the number of times a (sub)graph $g$ occurs in the set of graphs; 
$\mathit{support}(g_{1} \wedge g_{2})$ gives the number of graphs (or transactions) with both subgraphs and
$\mathit{support}(g_{1} \vee g_{2})$ gives the number of graphs with at least one of these subgraphs.
%In this measure we subtract $\mathit{support}(g_{1} \wedge g_{2})$ two times because the occurrences of 
%$g_{1}$ and $g_{2}$ occurring together is counted twice by $\mathit{support}(g_{1}) + \mathit{support}(g_{2})$. 
The numerator of the $\mathit{dist}$ measure computes the number of times the two graphs 
do not occur together in one graph of the dataset. We divide by $\mathit{support}(g_{1} \vee g_{2})$ 
to make the distance independent from the total occurrence,
thereby normalizing it. We can reformulate $\mathit{dist}$ in the following manner:

\begin{equation}\label{two} 
\mathit{dist}(g_{1}, g_{2}) = \frac{\mathit{support}(g_{1}) + \mathit{support}(g_{2}) - 2 \cdot \mathit{support}(g_{1} \wedge g_{2})}
{\mathit{support}(g_{1}) + \mathit{support}(g_{2}) - \mathit{support}(g_{1} \wedge g_{2})}
\end{equation}

\noindent In this way we do not need to separately compute $\mathit{support}(g_{1} \vee g_{2})$ 
by counting the number of times subgraphs occur in the graphs in the dataset, 
saving us the time needed to access this compressed dataset.

\begin{example}
Say fragment $A$ has a string 11100011 indicating in which molecule it occurs. 
So, from left to right, fragment $A$ occurs in the first 3 molecules and in the
last 2, where a 1 indicates that a fragment occurs and a 0 indicates a non-occurrence.
If fragment $B$ has string 01111000, we can see that either fragment $A$ or $B$ or both
($\mathit{support}(A \vee B)$) occur in 7 molecules $(5 + 4 - 2)$. Similarly,
$\mathit{support}(A \wedge B)=2$, and $\mathit{dist}(A,B)= 5/7$.\hfill$\Box$
\end{example}

The distance measure satisfies the usual requirements,
such as the triangular inequality.
Note that $0\leq\mathit{dist}(g_{1}, g_{2}) \leq 1$  and 
$\mathit{dist}(g_{1}, g_{2}) = 1 \Leftrightarrow \mathit{support}(g_{1} \wedge g_{2}) = 0$,
so $g_{1}$ and $g_{2}$ have no common transactions in this case.
If $\mathit{dist}(g_{1}, g_{2}) = 0$, both subgraphs occur in the same transactions,
but are not necessarily equal.

The distances are computed \emph{after} discovering all frequent subgraphs with \textsc{gSpan}. Obviously, 
while computing the support for the graphs not all frequent subgraphs are known and not all
distances can be computed while running \textsc{gSpan}.

\section{Preprocessing: Grouping}\label{group}

Possibly we will discover many frequent subgraphs, depending on the chosen minimal support. 
Next to having to store the supports of all frequent subgraphs, we will also have to store
the distance for all frequent subgraph combinations in order to decide clusters fast. 
If we have $n$ frequent subgraphs then storing the support for all $n(n-1)/2$ 
combinations might be too much. However many frequent subgraphs often are very similar 
in both structure and support. Furthermore, for these very similar frequent subgraphs 
there often exists a parent-child relation. 

Now we will propose a preprocessing step where we first group close subgraphs and we will
treat them as one point in our cluster model. This will reduce the number of points
in the cluster model and the number of distances that have to be stored and/or decided. This will
not only speed up the process of deciding the distance between groups or graphs, another benefit
is that the overview in the visualization will be improved because less of the same graphs are
in the 2-dimensional cluster model. Furthermore it will reduce exploration time for the expert,
because many of these redundant graphs are grouped. If an expert wants to view the occurrence
of a graph then (s)he can select just one of the group to be retrieved from the compressed set,
since their occurrence is almost equal.

%We call two graphs $g_{1}$ and $g_{2}$, where $g_{2}$ is a subgraph of $g_{1}$, \emph{close} if
%there is no other supergraph/subgraph combination $h_{1}$ and $h_{2}$ where 
%$\mathit{subgraph\_dist}(g_{1}, g_{2}) > \mathit{subgraph\_dist}(h_{1}, h_{2})$. 
The formula for the
distance between supergraph $g_2$ and subgraph $g_1$  originates from Equation~\ref{two}, where 
$\mathit{support}(g_{1} \wedge g_{2}) = \mathit{support}(g_{2})$:
\begin{eqnarray}\label{three} 
\mathit{dist}(g_{1}, g_{2})&=& \frac{\mathit{support}(g_{1}) + \mathit{support}(g_{2}) - 2 \cdot \mathit{support}(g_{2})}
{\mathit{support}(g_{1}) + \mathit{support}(g_{2}) - \mathit{support}(g_{2})}\nonumber\\
&=& \frac{\mathit{support}(g_{1}) - \mathit{support}(g_{2})}{\mathit{support}(g_{1})}\nonumber
\end{eqnarray}

%\noindent So the formula for $\mathit{subgraph\_dist}$ basically calculates $\mathit{dist}$ between super- and subgraph.

\begin{example}
If we take a fragment $A$ with occurrence string 11110011 and a fragment $B$
(where $A$ is a subgraph of $B$)
with string 11110000, then we of course see that fragment $A$ occurs at least in all
molecules where $B$ occurs. In this case
$\mathit{support}(A \wedge B) = \mathit{support}(B) = 4$.\hfill$\Box$
\end{example}

All information used to compute these distances can be retrieved from the lattice
information provided by the graph mining algorithm, when we focus on the subgraph-supergraph pairs. This information is needed by the graph 
mining algorithm to discover the frequent subgraphs and so the only extra calculating is done when
$\mathit{dist}$ does a search in this information.

Of course, many graphs have no parent-child relation and for this reason we define $\mathit{pregroup\_dist}$
in the following way:
\begin{equation}\label{vier} 
	\mathit{pregroup\_dist}(g_{1}, g_{2}) = \left\{ \begin{array}{l l} \mathit{dist}(g_{1}, g_{2}) & \quad \mbox{if $g_{2}$ is a supergraph of $g_{1}$}\\
	  & \quad \mbox{or $g_{1}$ is a supergraph of $g_{2}$}\\
	1 & \quad \mbox{otherwise}\\
	 \end{array} \right.
\end{equation}
Note that $\mathit{pregroup\_dist}(g_{1}, g_{2})<1$ if $g_1$ is a subgraph of $g_2$ and has non-zero support,
or the other way around.

Now we propose the \textsc{PreGroup} algorithm that will organize close
subgraphs/supergraphs into
groups. The algorithm is based on hierarchical clustering and because of this we need to define
how we decide the distance between clusters $C_{1} = \{g_{1}, g_{2}, \ldots , g_{n}\}$ and $C_{2} = \{h_{1}, h_{2}, \ldots , h_{m}\}$:
\begin{eqnarray}\label{vijf} 
\mathit{cluster\_dist}(C_{1}, C_{2}) = \left\{ \begin{array}{l l} \mathit{max}(\mathit{PG}) & \quad \mbox{if $\mathit{PG} \neq \emptyset$}\\
	-1 & \quad \mbox{otherwise}\\
	 \end{array} \right.\ \ \ \ \ \ \ \ \ \ \ \ \ \ \ \ \ \ \ \ \ \ \ \ \\
\mathit{PG} = \{\mathit{pregroup\_dist}(g, h)\; |\; g \in C_{1}, h \in C_{2}, \mathit{pregroup\_dist}(g, h) \neq 1\}\nonumber
\end{eqnarray}
Two clusters should not be merged if their graphs do not have a  supergraph-subgraph relation,
%or if there is no common transaction in which patterns exist. 
so we do not consider graphs where $\mathit{pregroup\_dist}(g, h) = 1$.
The value of $\mathit{cluster\_dist}$ is $-1$ if no maximal distance exists, and clusters will not be merged in the algorithm.

The outline of the algorithm is the following:

\vspace*{5mm}\hrule\vspace*{-1mm}
\begin{tabbing}
XX\=XX\=XX\=XX\=XX\=XX\kill
\>initialize $\cal{P}$ with sets of subgraphs of size 1 from the lattice \\
\>\textbf{while} $\cal{P}$ was changed or was initialized \\
\>\>Select $C_{1}$ and $C_{2}$ from $\cal{P}$ with minimal $\mathit{cluster\_dist}\;(C_{1}, C_{2})\geq 0$\\
\>\>\textbf{if} $\mathit{cluster\_dist}(C_{1}, C_{2}) \leq \mathit{maxdist}$ \textbf{then} \\
\>\>\>$\cal{P}$ = $\cal{P}$ $\cup\;\{C_{1} \cup C_{2}\}$ \\
\>\>\>Remove $C_{1}$ and $C_{2}$ from $\cal{P}$ \\
\end{tabbing}
\vspace*{-5mm}\hrule\vspace*{1mm}
\centerline{\textsc{PreGroup}}
\vspace*{1.2mm}
\hrule\vspace*{5mm}

\noindent The parameter $\mathit{maxdist}$ is a user-defined threshold giving the largest
distance allowed for two clusters to be joined.

\section{Relative Positioning of Groups}\label{relative}

% * SLIM SELECTEREN OP BASIS VAN HASH DIE GELIJKE STRUCTUUR EIGENSCHAPPEN SAMEN SORTEERD?

% * WAAROM IS HET ZO LANGSZAAM?

The information we need to store concerning the occurrence of subgraph patterns can be huge.
However, in some cases the user might want to have this information, e.g., in our working example
the scientist might want to closer investigate molecules (transactions) contain a specific pattern.

Interesting information for any user is to see how often the groups (clusters) of subgraphs occur
in the same transactions (graphs) within the dataset. Here we will visualize this co-occurrence
by positioning all groups randomly in a 2-dimensional area and adapting their position a number of times
with the formulas from \cite{KostersWA}. In our model for $\mathit{eucl\_dist}(C_{1},C_{2})$ 
we take the Euclidean distance between the 2-dimensional coordinates of the points corresponding with the
two groups (of frequent subgraphs) $C_1$ and $C_2$.

The graphs in a group occur in almost all the same transactions, depending on the chosen $\mathit{maxdist}$.
So, in order to speed up clustering, the distance between groups is assumed to be the distance between 
any of the points of the two groups. In our algorithm we choose to define the distance between groups as the distance
between a smallest graph of each of the two groups ($\mathit{size}$ gives the number of vertices):
for $g_{1} \in C_{1}$ and $g_{2} \in C_{2}$ with $size(g_{1}) = min(\{size(g)\;|\;g \in C_{1}\})$ 
and $size(g_{2}) = min(\{size(g)\;|\;g \in C_{2}\})$, we let $\mathit{group\_dist}(C_{1},C_{2}) = \mathit{dist}(g_{1}, g_{2})$.

The coordinates $(x_{C_{1}},y_{C_{1}})$ and $(x_{C_{2}},y_{C_{2}})$  of the points corresponding with $C_1$
and $C_2$ are adapted by applying
the following formulas:
\begin{enumerate}
	\item $x_{C_{1}} \leftarrow x_{C_{1}} - \alpha \cdot (\mathit{eucl\_dist}(C_{1},C_{2}) - \mathit{group\_dist}(C_{1},C_{2})) \cdot (x_{C_{1}} - x_{C_{2}})$
	\item $y_{C_{1}} \leftarrow y_{C_{1}} - \alpha \cdot (\mathit{eucl\_dist}(C_{1},C_{2}) - \mathit{group\_dist}(C_{1},C_{2})) \cdot (y_{C_{1}} - y_{C_{2}})$
	\item $x_{C_{2}} \leftarrow x_{C_{2}} + \alpha \cdot (\mathit{eucl\_dist}(C_{1},C_{2}) - \mathit{group\_dist}(C_{1},C_{2})) \cdot (x_{C_{1}} - x_{C_{2}})$
	\item $y_{C_{2}} \leftarrow y_{C_{2}} + \alpha \cdot (\mathit{eucl\_dist}(C_{1},C_{2}) - \mathit{group\_dist}(C_{1},C_{2})) \cdot (y_{C_{1}} - y_{C_{2}})$
\end{enumerate}
Here $\alpha$ ($0 \leq \alpha \leq 1$) is the user-defined learning rate. 

Starting with random coordinates for the groups, we will build a 2-dimensional model of relative positions between groups by randomly choosing two groups
$r$ times and applying the formulas.

\section{Results and Performance}\label{exps}

The experiments are organized such that we first show the cluster model to approximate the distances
correctly. 
Secondly through experiments we show the speed-up due to making groups first. We make use of
a dataset, the \emph{molecule dataset}, containing 4,069 molecules; from this we extracted a lattice containing the 1,229 most frequent subgraphs.

All experiments were performed on an Intel Pentium 4 64-bits 3.2 GHz machine with 3 GB memory. As operating system Debian
Linux 64-bits was used with kernel 2.6.8-12-em64t-p4.

\begin{figure}[!ht]
\begin{center}
\includegraphics[width=7.2cm]{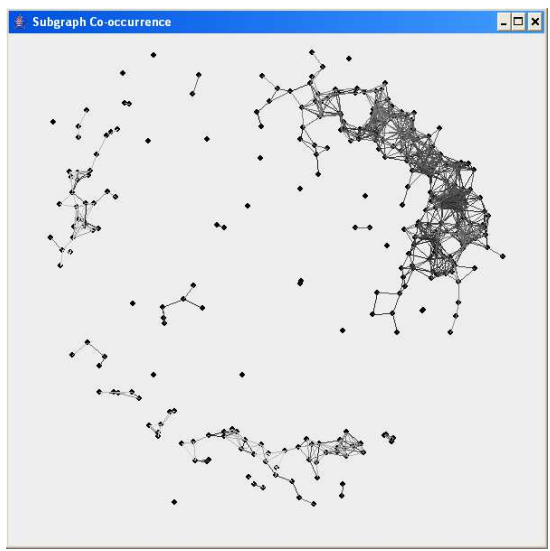}
\caption{Clusters for graphs in the molecule dataset built in 24.5 seconds, connecting points at distance 0.05 or lower
($\alpha = 0.1$, $\mathit{maxdist} = 0.1$, $r = 1,000,000$).}
\label{fig:clusterexample1}
\vspace*{-4mm}
\end{center}
\end{figure}

\begin{figure}[!ht]
\begin{center}
\includegraphics[width=7.2cm]{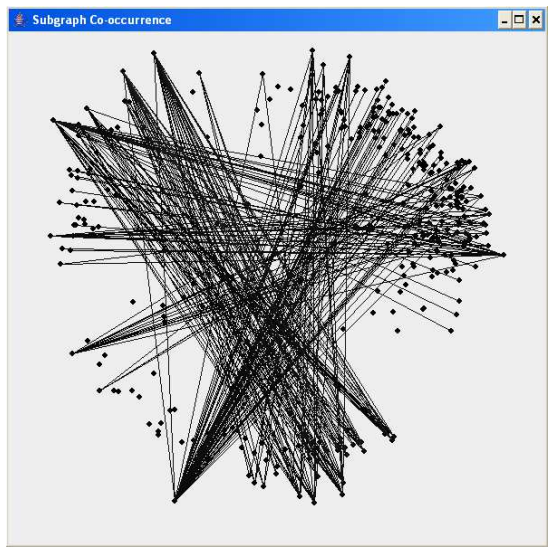}
\caption{Clusters of graphs in the molecule dataset built in 24.5 seconds, connecting points at distance 0.95 or higher
($\alpha = 0.1$, $\mathit{maxdist} = 0.1$, $r = 1,000,000$).}
\label{fig:clusterexample2}
\vspace*{-8mm}
\end{center}
\end{figure}

Figure \ref{fig:clusterexample1} shows how points, that represent subgraphs occurring in the same graphs
(molecules) of the dataset, cluster together. We made lines between points if their Euclidean distance
is $\leq 0.05$. The darker these lines the lower their actual distance and in this way one can see
gray clusters of close groups of subgraphs. Some groups are placed close but their actual distance is not
close (they are light grey). This is probably caused by the fact that these groups do not occur together with 
some specific other groups, so being far away from these other ones.

In Figure \ref{fig:clusterexample2} we make lines between points with a Euclidean distance
$\geq 0.95$. The darker these lines the higher their actual distance. The figure shows their actual
distance to be big also (the lines are black). Also Figure \ref{fig:clusterexample2} shows bundles
of lines going to one place. This probably is again caused by groups not occurring
together with the same other groups.

\begin{figure}[!ht]
\begin{center}
\includegraphics[width=7.0cm]{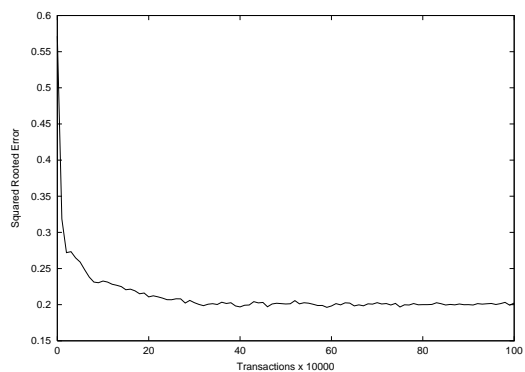}
\caption{Root squared error for distance given by the cluster model ($\alpha = 0.1$).}
\label{fig:error}
\vspace*{-10mm}
\end{center}
\end{figure}

The error for the cluster model decreases quickly, see Figure \ref{fig:error}. 
At some point it becomes very hard to reduce the error further.

In one experiment we assumed that the distances could not be stored in memory. In this experiment we first
clustered 1,229 patterns without grouping, taking 81 seconds. However, grouping reduced
the number of requests to the compressed occurrence data and because of this with grouping
clustering took 48 seconds ($\alpha = 0.1$, $r = 1,000,000$, $\mathit{maxdist} = 0.1$).

\begin{figure}[!ht]
\vspace*{-5mm}
\begin{center}
\includegraphics[width=7.0cm]{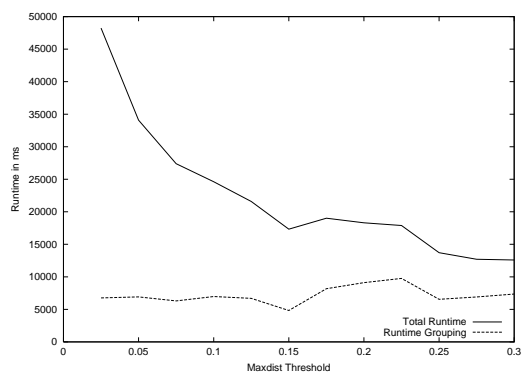}
\caption{Average runtime for the molecule dataset with varying $\mathit{maxdist}$ ($\alpha = 0.1$, $r = 1,000,000$).}
\label{fig:runtime}
\vspace*{-10mm}
\end{center}
\end{figure}

Our final experiment was done to show how the runtime is influenced by the $\mathit{maxdist}$ threshold
and how much the preprocessing step influences runtime. Here we assume the distances between
clusters can be stored in memory. In Figure~\ref{fig:runtime} the influence
on runtime is shown. The time for preprocessing appears to be more or less stable, but the total runtime drops
significantly.

\section{Conclusions and Future Work}

% * Conclusies:

Presenting data mining results to the user in an efficient way is important. 
In this paper we propose a preprocessing step for an existing method of clustering
and we apply this method to frequent subgraphs, which was not done before.

The model can be built faster with the clustering algorithm because of
the grouping of the subgraphs, the preprocessing step. The groups also
remove redundant points from the visualization that represents very similar
subgraph patterns. Finally the model enables the user to quickly select the
right subgraphs for which the user wants to investigate the graphs (or molecules) 
in which the frequent subgraphs occur.

In the future we want to take a closer look at grouping where the types of vertices
and edges and their corresponding weight also decide their group. Furthermore,
we want to investigate how we can compress occurrence more efficiently and
access it faster.

\bigskip 

\noindent \textbf{Acknowledgments}
This research is carried out within the Netherlands Organization for Scientific Research (NWO) MISTA Project (grant no. 612.066.304).
We thank Siegfried Nijssen for his implementation of \textsc{gSpan}.

\end{document}